\documentclass[a4paper,fleqn]{cas-dc}

\usepackage[authoryear,longnamesfirst]{natbib}

\usepackage{algorithm}
\usepackage{algpseudocode}
\usepackage{listings}
\usepackage{xcolor}
\usepackage{graphicx}
\usepackage{booktabs}
\usepackage{makecell}
\usepackage{placeins}
\usepackage{multirow}
\usepackage{caption}
\usepackage{adjustbox}
\usepackage{bm}
\usepackage{float}
\usepackage{amssymb}
\usepackage{amsmath}
\def\tsc#1{\csdef{#1}{\textsc{\lowercase{#1}}\xspace}}
\tsc{WGM}
\tsc{QE}

\begin{document}
\let\WriteBookmarks\relax

\shorttitle{}    

\shortauthors{}  

\title [mode = title]{High-Resolution Underwater Camouflaged Object Detection: GBU-UCOD Dataset and Topology-Aware and Frequency-Decoupled Networks}  

\tnotemark[1] 

\tnotetext[1]{} 

\author[1]{Wenji Wu}
\fnmark[$\dagger$]        
\credit{Data collection, Methodology, Writing - Original draft preparation, Editing}

\author[2]{Shuo Ye}
\fnmark[$\dagger$]
\credit{Supervision, Review}

\author[2]{Yiyu Liu}
\fnmark[$\dagger$]
\credit{Data curation}

\author[2]{Jiguang He}
\credit{Review}

\author[1]{Zhuo Wang}
\cormark[1]
\ead{wangzhuo@hrbeu.edu.cn}
\credit{Supervision, Review}

\author[2]{Zitong Yu}
\cormark[1]
\ead{zitong.yu@gbu.edu.cn}
\credit{Project administration, Supervision, Review}



\affiliation[1]{%
  organization={College of Shipbuilding Engineering, Harbin Engineering University},
  city={Harbin},
  postcode={150001},
  country={China}
}

\affiliation[2]{%
  organization={School of Information Science and Technology, Great Bay University},
  city={Dongguan},
  postcode={523000},
  country={China}
}

\cortext[cor1]{Corresponding authors: Zhuo Wang and Zitong Yu (Project Leader).}

\fntext[1]{$\dagger$ Wenji Wu, Shuo Ye, and Yiyu Liu contribute equally to this work.}

\fntext[2]{This work was performed while Wenji Wu was an intern at Great Bay University.}


\begin{abstract}
Underwater Camouflaged Object Detection (UCOD) is a challenging task due to the extreme visual similarity between targets and backgrounds across varying marine depths. Existing methods often struggle with topological fragmentation of slender creatures in the deep sea and the subtle feature extraction of transparent organisms. In this paper, we propose DeepTopo-Net, a novel framework that integrates topology-aware modeling with frequency-decoupled perception. To address physical degradation, we design the Water-Conditioned Adaptive Perceptor (WCAP), which employs Riemannian metric tensors to dynamically deform convolutional sampling fields. Furthermore, the Abyssal-Topology Refinement Module (ATRM) is developed to maintain the structural connectivity of spindly targets through skeletal priors. Specifically, we first introduce GBU-UCOD, the first high-resolution (2K) benchmark tailored for marine vertical zonation, filling the data gap for hadal and abyssal zones. Extensive experiments on MAS3K, RMAS, and our proposed GBU-UCOD datasets demonstrate that DeepTopo-Net achieves state-of-the-art performance, particularly in preserving the morphological integrity of complex underwater patterns. The datasets and codes will be released at https://github.com/Wuwenji18/GBU-UCOD.
\end{abstract}



\maketitle

\section{Introduction}
Underwater Camouflaged Object Detection (UCOD) aims to identify and segment marine organisms that are visually integrated into their aquatic environments \cite{fan2020camouflage}. As a fundamental component of underwater computer vision, robust UCOD capabilities are indispensable for enabling Autonomous Underwater Vehicles (AUVs) to execute intelligent perception tasks, such as ecological monitoring and automated marine resource exploration \cite{jmse11061119}. However, reliance on simple texture cues often fails due to the natural light loss in water, which necessitates specialized correction strategies \cite{li2017hybrid,zhang2025mact}. In the deep sea, this challenge is intensified by severe scattering and the complex shapes of marine life, making detection significantly harder \cite{zhang2021underwaterPRL}. Without clear geometric guides to handle such physical quality loss, current models often produce broken predictions for thin limbs or miss transparent creatures entirely. Moreover, since existing models are mainly trained on shallow-water or land data, they struggle to work effectively across different ocean depths. A major barrier to progress is the lack of datasets specifically designed for deep-sea camouflaged objects.

\begin{figure*}[hbpt]
    \centering
    \includegraphics[width=0.95\textwidth]{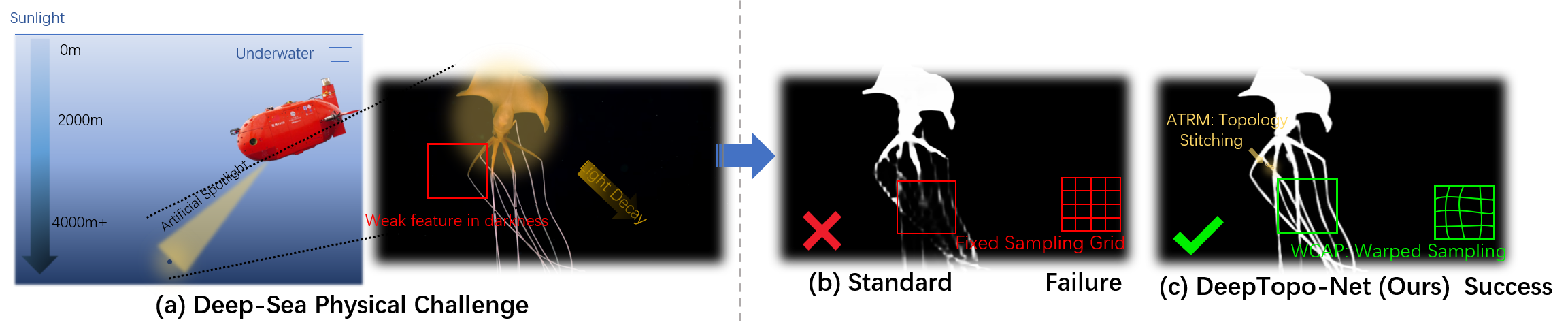}
    \caption{\textbf{Motivation.} \textbf{(a) Challenge:} Artificial spotlights obscure fine topology (e.g., tentacles) in dark zones. \textbf{(b) Baseline Failure:} Rigid sampling grids miss these faint features, causing fragmentation. \textbf{(c) Our Success:} Our DeepTopo-Net recovers the complete structure using WCAP (warped sampling) and ATRM (topology stitching).}
    \label{fig:motivation}
\end{figure*}

To address these limitations, we propose DeepTopo-Net, a unified framework that integrates topology-aware modeling with frequency-decoupled perception to handle the complex conditions of deep-sea environments. To address severe light degradation and inconsistent attenuation across different depths, we design the Water-Conditioned Adaptive Perceptor (WCAP), utilizes Riemannian metric tensors to dynamically warp the convolutional sampling field. This mechanism enables the network to adaptively compensate for non-uniform light attenuation, significantly enhancing the observability of transparent and low-contrast targets to improve segmentation accuracy. Second, to resolve the fragmentation of slender structures, we introduce the Abyssal-Topology Refinement Module (ATRM) to enforce skeletal connectivity. By incorporating explicit geometric priors, this module effectively "stitches" disconnected segments of slender limbs, ensuring the morphological integrity of the predicted masks. Finally, to advance the field of deep-sea COD, we present GBU-UCOD, the first high-resolution benchmark tailored for marine vertical zonation. Unlike existing datasets restricted to shallow waters, Our dataset captures extreme biological samples from hadal and abyssal zones, providing a critical data foundation for robust deep-sea research. Extensive experiments on MAS3K \cite{li2021marine}, RMAS \cite{10113781}, and the proposed GBU-UCOD datasets demonstrate that DeepTopo-Net consistently achieves state-of-the-art performance. In particular, our method significantly outperforms existing models in preserving the morphological integrity of complex deep-sea targets, establishing a new baseline for high-resolution underwater perception.

The main contributions of this work are summarized as follows:
\begin{itemize}
    \item \textbf{We construct GBU-UCOD, the first high-resolution benchmark specifically designed for marine vertical zonation.} By incorporating diverse samples from the hadal and abyssal zones, this dataset has contributed to  deep-sea research and facilitates the study of extreme biological adaptations.
    \item \textbf{We propose the Water-Conditioned Adaptive Perceptor (WCAP) to counteract non-uniform optical degradation.} Utilizing Riemannian metric tensors to dynamically warp the convolutional sampling field, this module effectively compensates for light attenuation and significantly enhances the observability of transparent targets.
    \item \textbf{We introduce the Abyssal-Topology Refinement Module (ATRM) to address topological fragmentation.} By integrating explicit skeletal connectivity priors, this mechanism effectively ``stitches'' disconnected segments of slender limbs, ensuring the morphological integrity of complex deep-sea organisms.
\end{itemize}

\section{Related Work}
\label{sec:related}

\subsection{Underwater Camouflaged Object Detection}
The objective of COD is to identify objects visually embedded in their surroundings. In terrestrial scenarios, deep learning-based methods have established a ``coarse-to-fine'' paradigm \cite{fan2020camouflage,fan2021concealed}. To distinguish targets from clutter, recent approaches rely on exploring auxiliary modalities like polarization \cite{wang2023polarization} or mining fine-grained cues: HitNet \cite{hu2023high} enhances high-resolution texture details via iterative feedback, while ZoomNet \cite{pang2022zoom} employs mixed-scale zooming to capture ambiguous boundaries. Similarly, AGNet \cite{zhou2022agnetPRL} utilized attention guidance to effectively locate concealed targets. Additionally, structural dependencies are often modeled through explicit boundary guidance \cite{sun2022boundary} or mutual graph learning \cite{zhai2021mutual}.

However, directly transferring these terrestrial models to the marine environment yields suboptimal results, as the fine-grained texture cues they rely on are often obliterated by severe light attenuation and scattering. To address this, the focus has shifted towards Underwater COD (UCOD). Benchmarks like MAS3K and RMAS have moved beyond general object detection to segmenting camouflaged marine animals, proposing specialized baselines like MASNet \cite{10113781}. 

Despite these contributions, existing UCOD research exhibits a strong bias towards shallow epipelagic waters, where natural sunlight is abundant. They largely overlook the abyssal and hadal zones, which constitute the majority of the ocean's volume. In these hostile deep-sea environments, organisms evolve unique morphological features—such as extreme transparency or fragile elongated appendages—to survive in high-pressure, light-deprived conditions. Current models trained on shallow-water data struggle to recognize these distinct topological patterns, and the absence of a high-resolution benchmark covering marine vertical zonation remains a primary bottleneck.

\begin{figure*}[hbpt]
    \centering
    \includegraphics[width=\linewidth]{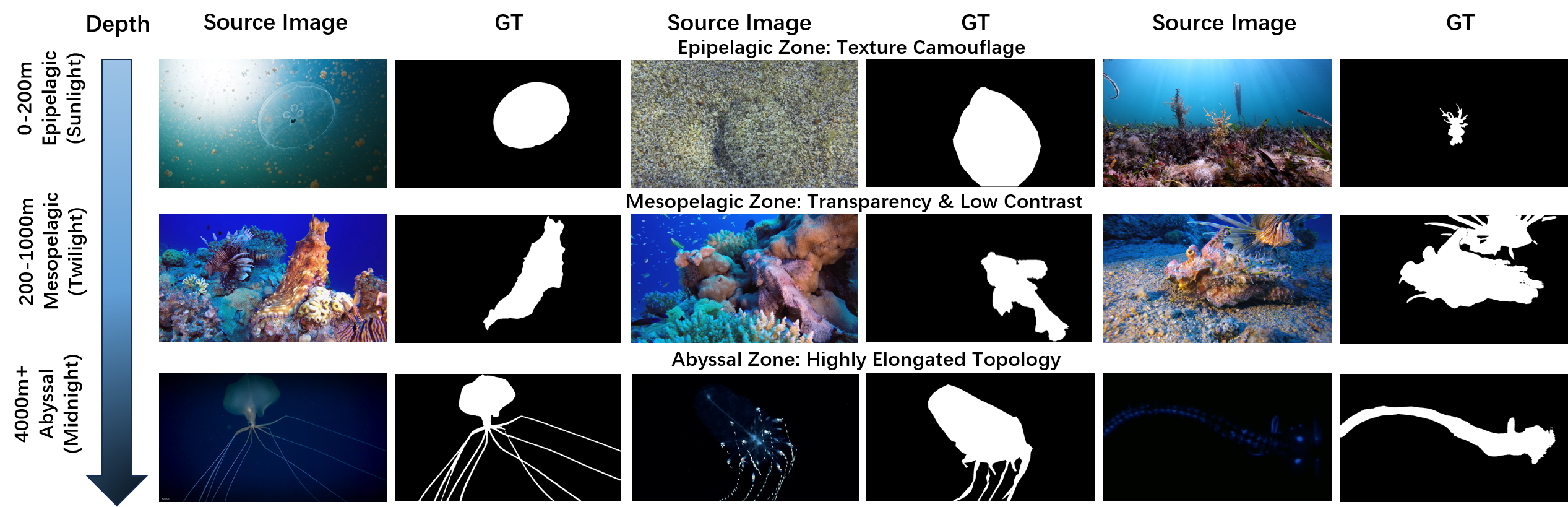}
    \caption{Vertical zonation overview illustrating the challenges across different depth strata in the proposed GBU-UCOD benchmark.}
    \label{fig:dataset-overview}
\end{figure*}

\subsection{Riemannian Geometry in Computer Vision}
Early deep manifold learning frameworks, such as SPDNet \cite{huang2017riemannian}, introduced Riemannian-specific layers to learn discriminative SPD matrices for image classification. Subsequently, Brooks et al. \cite{brooks2019riemannian} proposed Riemannian Batch Normalization to stabilize the training of deep manifold networks. More recently, the field has expanded to complex optimization and dense prediction tasks. Gao et al. \cite{gao2020learning} introduced a learnable optimizer to automatically navigate the SPD manifold geometry. Chen et al. \cite{chen2023riemannian} developed a multi-scale submanifold network to explicitly mine local geometric information for fine-grained recognition. Furthermore, recent works like GeloVec \cite{kriuk2025gelovechigherdimensionalgeometric} have applied Riemannian smoothing to attention mechanisms to enhance segmentation stability. Different from prior works that mostly use Riemannian geometry for statistical alignment or feature smoothing, we introduce it into dense pixel-level feature sampling. By constructing a learnable Riemannian metric tensor to model the local anisotropy of deep-sea optical distortion, our WCAP module enables the network to adaptively perceive the warped visual space, effectively compensating for the non-uniform light propagation in the abyss, akin to optical property-based restoration methods \cite{wang2023polarization}.


\subsection{Topology-Aware and Adaptive Learning}
Beyond standard segmentation, preserving the structural integrity of thin objects remains a frontier challenge. Recently, Skeleton Recall Loss \cite{kirchhoff2024skeleton} and SATLoss \cite{wen2025topology} introduced connectivity-conserving strategies to penalize topological breaks. In the realm of adaptive perception, attention mechanisms like SENet \cite{hao2025simple} and APGNet \cite{huang2025apgnet} have been widely adopted to recalibrate feature responses, while recent methods like FE-UNet \cite{huo2025frequencydomainenhancedunet} and FADC \cite{chen2024frequency} exploit frequency components to enhance detailed feature representation. Furthermore, the emergence of foundation models has reshaped the landscape. Hybrid architectures such as H2Former \cite{he2023h2former} and SAM-based adaptations, including Dual-SAM \cite{zhang2024fantastic}, MAS-SAM \cite{wang2024massam}, and SAM2-UNet \cite{xiong2026sam2} which have pushed performance boundaries by leveraging massive pre-trained priors.


However, a critical gap remains in the deep-sea domain. While these advanced baselines possess powerful semantic abstraction capabilities, they often lack the explicit geometric constraints required to recover degraded deep-sea structures. Deep-sea organisms present a dual challenge: their topological complexity is compounded by physical visual degradation. Direct application of general attention or hybrid transformers often fails because raw underwater features are too weak to support fine-grained recovery. Therefore, a unified framework that synergizes physically-grounded enhancement with topology-specific refinement is needed.

\FloatBarrier
\section{GBU-UCOD Dataset}
\label{sec:dataset}

Existing underwater datasets are mostly collected in shallow waters, lacking the diversity of vertical zonation. To address this, we construct a new benchmark named GBU-UCOD.


\noindent {\bf Vertical Zonation and Data Collection.}
As illustrated in Fig. \ref{fig:dataset-overview}, GBU-UCOD covers a wide depth range from 0m to over 4000m, featuring high-resolution (2K) source images. Unlike previous datasets, we explicitly categorize samples into three vertical zones based on light penetration:
\begin{enumerate}
    \item \textbf{Epipelagic Zone (Sunlight, 0-200m):} Characterized by bright natural light but complex texture camouflage (e.g., flounder blending into sand).
    \item \textbf{Mesopelagic Zone (Twilight, 200-1000m):} Dominated by dim blue light, featuring low-contrast organisms like octopuses hiding in corals.
    \item \textbf{Abyssal Zone (Midnight, 4000m+):} A completely dark abyssal environment illuminated only by non-uniform artificial light, containing unique organisms with highly elongated and fragile topologies.
\end{enumerate}

\noindent {\bf Dataset Statistics.}
The dataset contains 1,052 image-mask pairs. As shown in the chord diagram, the largest category is Deep-sea organism (336 samples), followed by Camouflaged octopus (178). 
Rather than enforcing a balanced class distribution, this dataset emphasizes ecologically realistic and visually challenging deep-sea scenarios, where camouflaged and morphologically fragile organisms are dominant.

Such a distribution reflects real-world underwater observation conditions and highlights the limitations of existing shallow-water–biased benchmarks, thereby providing a more rigorous testbed for evaluating the robustness and generalization ability of UCOD methods.

\noindent {\bf High-Quality Annotation.}
As demonstrated in the ``Ground Truth'' columns of Fig. \ref{fig:dataset-overview}, we adopt a rigorous pixel-level annotation strategy. Special attention is paid to the topological integrity of abyssal creatures, ensuring that extremely slender limbs (e.g., the tentacles in the bottom row) are accurately labeled without disconnection.

\begin{figure*}[!t] 
	\centering
	\includegraphics[width=0.98\textwidth]{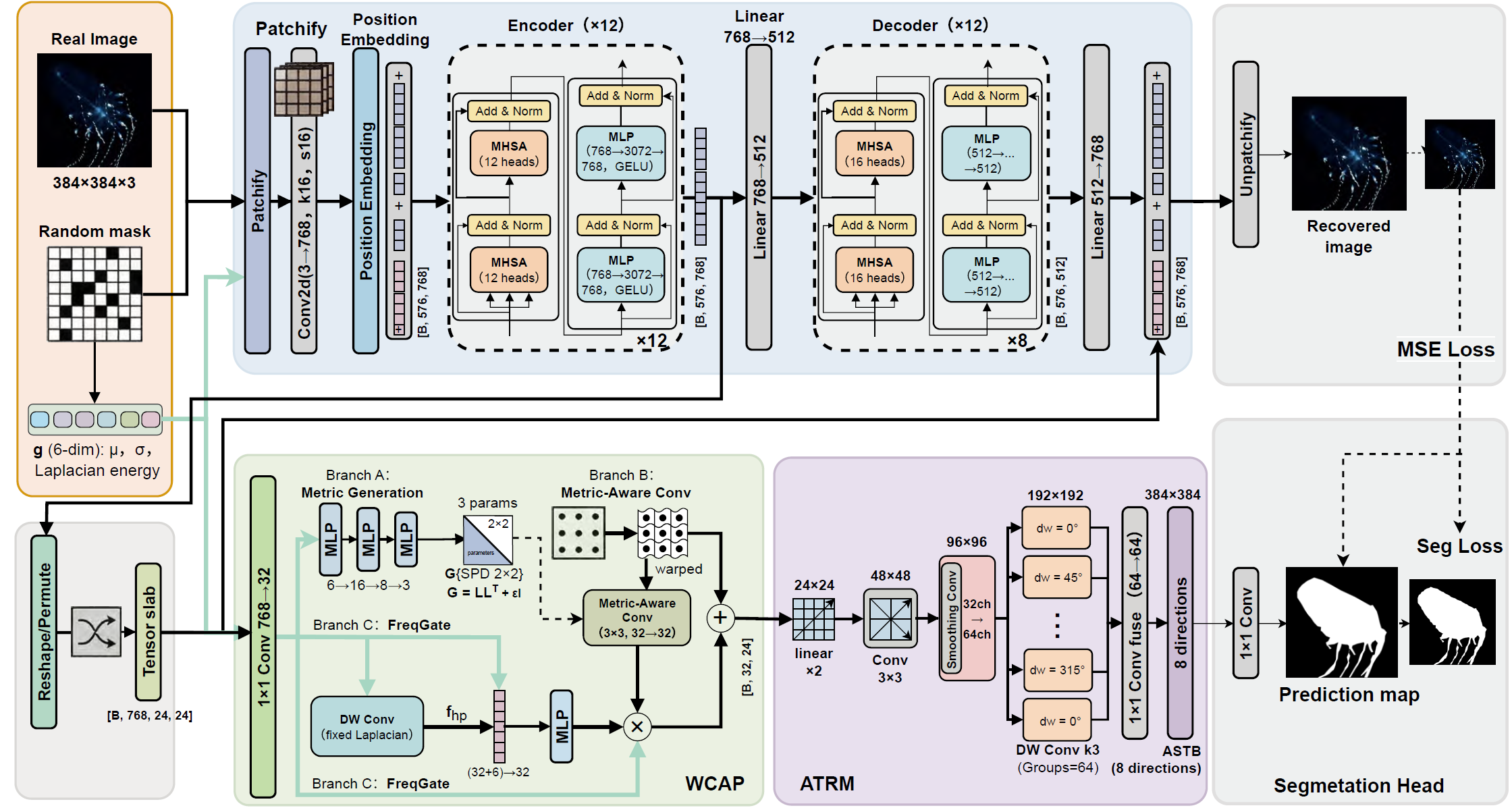}
	\caption{The overall architecture of DeepTopo-Net. The framework utilizes a MAE-ViT backbone for semantic reconstruction. The WCAP module compensates for physical degradation using Riemannian metric warping, while the ATRM utilizes directional geometric mining and skeletal constraints to ensure topological connectivity of slender targets.}
	\label{fig:arch} 
\end{figure*}

\FloatBarrier
\section{Methodology}
\label{sec:method}

\subsection{Overall Framework}
Fig. \ref{fig:arch} illustrates the overall training framework of the proposed DeepTopo-Net. DeepTopo-Net is a unified multi-task model built upon an asymmetric ViT-based Masked Autoencoder (MAE) structure \cite{he2022masked,huang2025multi,hao2025simple}. It can be seen that we use both image reconstruction and binary segmentation tasks as our training targets, where image reconstruction acts as the auxiliary task to enhance feature robustness \cite{hao2025simple}. Specifically, the input image is first serialized into tokens where a substantial portion is randomly masked, and the encoder operates on the remaining visible tokens to model global contexts. For the auxiliary branch, a lightweight decoder restores the removed tokens to reconstruct the original image. For the main segmentation branch, the latent features are first bridged by the Water-Conditioned Adaptive Perceptor (WCAP), which integrates global statistics to compensate for optical degradation. After that, the Abyssal-Topology Refinement Module (ATRM) serves as a topology-aware decoder to refine skeletal connectivity. At last the processed features are restored to the original resolution, and the required single-channel prediction map is obtained through a $1\times1$ convolution operation. The formalized execution logic of DeepTopo-Net is detailed in Algorithm \ref{alg:main}.

\subsection{Water-Conditioned Adaptive Perceptor}
\label{subsec:wcap}
Normal convolutional neural networks assume that feature distributions reside in a uniform Euclidean space, an assumption that fails under the non-uniform light propagation of the abyss. WCAP addresses this by redefining the sampling process within a Riemannian metric manifold.

\noindent {\bf Riemannian Metric Tensor Construction.}
We characterize the local geometric distortion using a metric tensor $\mathbf{G} \in \mathbb{R}^{2 \times 2}$, the global descriptor $\bm{g}$ (6-dim) is projected by an MLP to predict 3 specific parameters. These parameters populate a lower-triangular matrix $\mathbf{L}(\bm{g})$, constructing the SPD metric tensor via Cholesky decomposition:
\begin{equation}
\mathbf{G}(\bm{g}) = \mathbf{L}(\bm{g})\mathbf{L}(\bm{g})^T + \epsilon \mathbf{I}.
\end{equation}

\noindent {\bf Metric-Aware Manifold Sampling.}
To achieve physical compensation, the standard grid is replaced by a warped sampling field (Branch B). Driven by the metric $\mathbf{G}$, the receptive field adaptively stretches to compensate for attenuation. The warped feature map $f_{warped}$ is obtained by sampling the input $f_{in}$ based on the metric-induced distance:
\begin{equation}
d_{\mathbf{G}}(\Delta p) = \sqrt{\Delta p^T \mathbf{G}(\bm{g}) \Delta p}.
\end{equation}

\noindent {\bf Frequency-Gated Saliency Enhancement.}
Simultaneously, Branch C employs a fixed Laplacian operator to extract high-frequency structural cues $f_{hp}$. To integrate global context, we concatenate the pooled high-frequency features with $\bm{g}$ (denoted as $32+6$ in Fig. \ref{fig:arch}). The final output $f_{out}$ is the fusion of the warped stream and the gated frequency stream:
\begin{equation}
f_{out} = f_{warped} + \underbrace{\text{Sigmoid}\left( \text{MLP}(\text{GAP}(f_{hp}) \otimes \bm{g}) \right) \cdot f_{hp}}_{\text{FreqGate Response}},
\end{equation}
where $\otimes$ denotes concatenation. This residual design ensures that geometric correction and edge enhancement are synergistically combined.

\subsection{Abyssal-Topology Refinement Module}
\label{subsec:atrm}
Standard decoders (e.g., in U-Net\cite{xiong2026sam2}) typically rely on simple bilinear upsampling or deconvolution. While effective for large objects, these isotropic operations often blur or fracture the extremely slender limbs of deep-sea organisms, leading to disconnected predictions. To address this, we propose the Abyssal-Topology Refinement Module (ATRM), which acts as a topology-aware decoder to explicitly restore skeletal connectivity.

\noindent {\bf Progressive Upsampling and Smoothing.}
ATRM first restores the spatial resolution from the bottleneck ($24 \times 24$) to the fine scale ($384 \times 384$). Unlike direct upsampling, we employ a ``Smoothing Envelope'' strategy. After each bilinear interpolation step, a $3 \times 3$ convolution with BatchNorm and ReLU is applied. This stabilizes the feature boundaries before they are fed into the directional filter.

\noindent {\bf Directional Skeletal Filtering (ASTB).}
The core component is the Anisotropic Structural Tensor Block (ASTB), designed to stitch fragmented limbs. Since marine appendages can be oriented in any direction, a single fixed kernel is insufficient. We devise a bank of directional filters covering 8 discrete orientations:
\begin{equation}
\Theta = \{0^\circ, 45^\circ, 90^\circ, \dots, 315^\circ\}.
\end{equation}
For each angle $\theta \in \Theta$, we apply a specific Depth-Wise convolution kernel $K_\theta$ aligned with that direction. This allows the network to ``scan'' along potential limb paths.

\noindent {\bf Directional Fusion.}
The responses from all 8 directions are then aggregated to form a complete topological map. We employ a $1 \times 1$ fusion convolution to merge directional cues:
\begin{equation}
f_{refined} = \text{Conv}_{1 \times 1} \left( \text{Concat}\big( [ \text{Conv}_{\theta}(f_{in}) \mid \forall \theta \in \Theta ] \big) \right).
\end{equation}
By explicitly integrating responses from all geometric orientations, ATRM ensures that even the thinnest connections are preserved in the final prediction.

\subsection{Loss Function}
The DeepTopo-Net is optimized via a multi-task loss $\mathcal{L}_{total}$, which combines an auxiliary reconstruction objective with a primary segmentation target.

\noindent {\bf Reconstruction Loss.} We follow the MAE paradigm \cite{he2022masked} to supervise the image recovery task. The Mean Squared Error (MSE) is calculated only for the masked patches to encourage the encoder to learn robust structural features:
\begin{equation}
\mathcal{L}_{rec} = \text{MSE}(I_{rec}, I_{gt}),
\end{equation}
where $I_{rec}$ and $I_{gt}$ represent the recovered and original images, respectively.

\noindent {\bf Segmentation Loss.} To prevent the fragmentation of slender limbs in deep-sea organisms, we employ a Dynamic Weighted (DW) Loss based on BCE and IoU. The core strategy is to prioritize the stitching areas of the biological skeleton. Specifically, we define a weight $\alpha$ for the target region that adaptively scales with the object size:
\begin{equation}
\alpha = l \cdot \frac{S_{img}}{S_{obj}},
\end{equation}
where $S_{img}$ is the image area and $S_{obj}$ is the target area. By applying this weight to the pixel-wise matrix $W$, the network is forced to focus on thin appendages and potential topological breaks. The final segmentation loss is formulated as:
\begin{equation}
\mathcal{L}_{seg} = W \odot (\text{BCE}(I_{pred}, I_{gt}) + \text{IoU}(I_{pred}, I_{gt})),
\end{equation}
where $\odot$ denotes pixel-wise multiplication and averaging. This weighting mechanism ensures that the ATRM effectively ``stitches'' broken segments by penalizing skeletal discontinuities.

\noindent {\bf Total Loss.} The overall training objective is a weighted summation of the two tasks:
\begin{equation}
\mathcal{L}_{total} = \lambda \cdot \mathcal{L}_{rec} + (1 - \lambda) \cdot \mathcal{L}_{seg},
\end{equation}
where $\lambda$ is a hyperparameter set to 0.1 to balance the auxiliary and main tasks.

\begin{algorithm}[t]
\caption{Training Procedure of DeepTopo-Net}
\label{alg:main}
\begin{algorithmic}[1]
\Require Training Batch $\mathbf{I}$, Ground Truth $\mathbf{Y}_{gt}$
\Ensure Optimized Parameters $\Theta$

\Statex \textbf{\textit{Phase 1: Water-Conditioned Adaptive Perception (WCAP)}}
\State $\mathcal{F} \gets \textsc{Encoder}(\mathbf{I})$; \quad $\bm{g} \gets \textsc{GAP}(\mathcal{F}_{last})$
\State $\mathbf{G} \gets \textsc{Metric}(\bm{g})$; \quad $\mathcal{F}_{warped} \gets \textsc{Warp}(\mathcal{F}, \mathbf{G})$
\State $\mathcal{F}_{enh} \gets \mathcal{F}_{warped} + \textsc{FreqGate}(\mathcal{F}_{warped}, \bm{g})$

\Statex \textbf{\textit{Phase 2: Abyssal-Topology Refinement (ATRM)}}
\State $f_{dec} \gets \textsc{Decoder}(\mathcal{F}_{enh})$
\State $\mathcal{R}_{\Theta} \gets \{ \textsc{Conv}_{\theta}(f_{dec}) \mid \theta \in \{0^\circ \dots 315^\circ\} \}$
\State $\mathbf{Y}_{pred} \gets \textsc{ASTB}(f_{dec}, \mathcal{R}_{\Theta})$

\Statex \textbf{\textit{Phase 3: Joint Optimization}}
\State $\mathcal{L}_{total} \gets \mathcal{L}_{\text{seg}}(\mathbf{Y}_{pred}, \mathbf{Y}_{gt}) + \lambda \cdot \mathcal{L}_{\text{rec}}$
\State $\Theta \gets \textsc{AdamW}(\Theta, \nabla \mathcal{L}_{\text{total}})$

\end{algorithmic}
\end{algorithm}

\FloatBarrier
\section{Experiments}
\label{sec:experiments}

\subsection{Experimental Setup}
\label{subsec:setup}
This section describes the experimental setup used to evaluate DeepTopo-Net, including the datasets, evaluation metrics, and implementation details.

\noindent {\bf Datasets.} We conduct experiments on three used underwater benchmarks: MAS3K \cite{li2021marine}, RMAS \cite{10113781}, and our newly proposed GBU-UCOD dataset. Following standard practices, we use the training sets of these benchmarks to train our model and evaluate its performance on the corresponding testing sets.

\noindent {\bf Metrics.} Following, five standard metrics are reported to evaluate the segmentation performance: structure measure ($S_\alpha$), weighted F-measure ($F_\beta^w$), mean E-measure ($mE_\phi$), mean Intersection over Union (mIoU), and mean absolute error (MAE). These metrics provide a comprehensive assessment of both pixel-level accuracy and structural integrity. 

\noindent {\bf Implementation details.} DeepTopo-Net is implemented in PyTorch on an NVIDIA RTX 4090 GPU and optimized using the AdamW optimizer. For the transformer-based encoder, we set the dimension to 768, with 12 heads and a depth of 12. For the decoder, the dimension is 512, with 16 heads and a depth of 8. We initialize the model using weights from a pre-trained MAE. All input images are resized to $384 \times 384$. During training, we set the masking ratio to 5\% for the auxiliary reconstruction task. The learning rate is initialized to $5 \times 10^{-5}$.

\begin{table*}[t]
\centering
\caption{Quantitative comparison on MAS3K, RMAS, and GBU-UCOD benchmarks. Best and second-best results are in \textbf{bold} and \underline{underline}, respectively. $\uparrow$ indicates the higher the score the better, and $\downarrow$ indicates the lower the better.}
\label{tab:sota}
\resizebox{\textwidth}{!}{
\begin{tabular}{l|c|ccccc|ccccc|ccccc}
\hline
\textbf{Method} & \textbf{Year} & \multicolumn{5}{c|}{\textbf{MAS3K}} & \multicolumn{5}{c|}{\textbf{RMAS}} & \multicolumn{5}{c}{\textbf{GBU-UCOD}} \\ \cline{3-17} 
 &  & mIoU$\uparrow$ & $S_\alpha\uparrow$ & $F_\beta^w\uparrow$ & $mE_\phi\uparrow$ & MAE$\downarrow$ & mIoU$\uparrow$ & $S_\alpha\uparrow$ & $F_\beta^w\uparrow$ & $mE_\phi\uparrow$ & MAE$\downarrow$ & mIoU$\uparrow$ & $S_\alpha\uparrow$ & $F_\beta^w\uparrow$ & $mE_\phi\uparrow$ & MAE$\downarrow$ \\ \hline
SINet & 2020 & 0.682 & 0.785 & 0.724 & 0.842 & 0.058 & 0.675 & 0.792 & 0.731 & 0.865 & 0.049 & 0.695 & 0.768 & 0.715 & 0.854 & 0.062 \\
PFNet & 2021 & 0.705 & 0.808 & 0.752 & 0.865 & 0.052 & 0.692 & 0.815 & 0.760 & 0.882 & 0.044 & 0.718 & 0.792 & 0.748 & 0.875 & 0.055 \\
MASNet & 2024 & 0.742 & 0.864 & 0.788 & 0.906 & 0.032 & 0.731 & 0.862 & 0.801 & 0.920 & 0.024 & 0.755 & 0.858 & 0.802 & 0.905 & 0.039 \\
H2Former & 2024 & 0.748 & 0.865 & 0.810 & 0.925 & 0.028 & 0.717 & 0.844 & 0.799 & 0.931 & 0.023 & 0.761 & 0.863 & 0.808 & 0.909 & 0.037 \\
Dual-SAM & 2024 & 0.799 & 0.884 & 0.838 & 0.933 & 0.023 & 0.735 & 0.860 & 0.812 & 0.944 & 0.022 & 0.782 & 0.871 & 0.825 & 0.912 & 0.035 \\
MAS-SAM & 2024 & 0.788 & 0.887 & 0.840 & 0.938 & 0.025 & 0.742 & 0.865 & 0.819 & \textbf{0.948} & \textbf{0.021} & 0.795 & 0.882 & 0.836 & 0.928 & 0.032 \\
SENet & 2025 & 0.780 & 0.877 & \underline{0.854} & 0.927 & 0.028 & 0.720 & 0.863 & \underline{0.857} & \underline{0.945} & 0.023 & 0.790 & 0.870 & 0.810 & 0.932 & 0.036 \\
SAM2-Unet & 2025 & 0.799 & \underline{0.903} & 0.848 & \textbf{0.943} & \textbf{0.021} & 0.738 & \underline{0.874} & 0.810 & 0.944 & 0.022 & \underline{0.817} & \underline{0.909} & \underline{0.852} & \textbf{0.944} & \underline{0.030} \\
APGNet & 2025 & 0.767 & 0.891 & 0.822 & 0.929 & 0.026 & 0.742 & 0.872 & 0.822 & 0.936 & \textbf{0.021} & 0.765 & 0.868 & 0.814 & 0.921 & 0.038 \\
FE-Unet & 2025 & \underline{0.815} & 0.900 & 0.848 & 0.928 & 0.022 & \textbf{0.758} & \underline{0.874} & 0.811 & 0.938 & \textbf{0.021} & 0.802 & 0.885 & 0.828 & 0.925 & 0.032 \\ 
\textbf{Ours} & - & \textbf{0.804} & \textbf{0.905} & \textbf{0.864} & \underline{0.941} & \textbf{0.021} & \underline{0.742} & \textbf{0.875} & \textbf{0.826} & 0.939 & \textbf{0.021} & \textbf{0.829} & \textbf{0.918} & \textbf{0.859} & \underline{0.939} & \textbf{0.028} \\ \hline
\end{tabular}
}
\end{table*}

\begin{figure*}[t]
	\centering
	\includegraphics[width=0.99\textwidth]{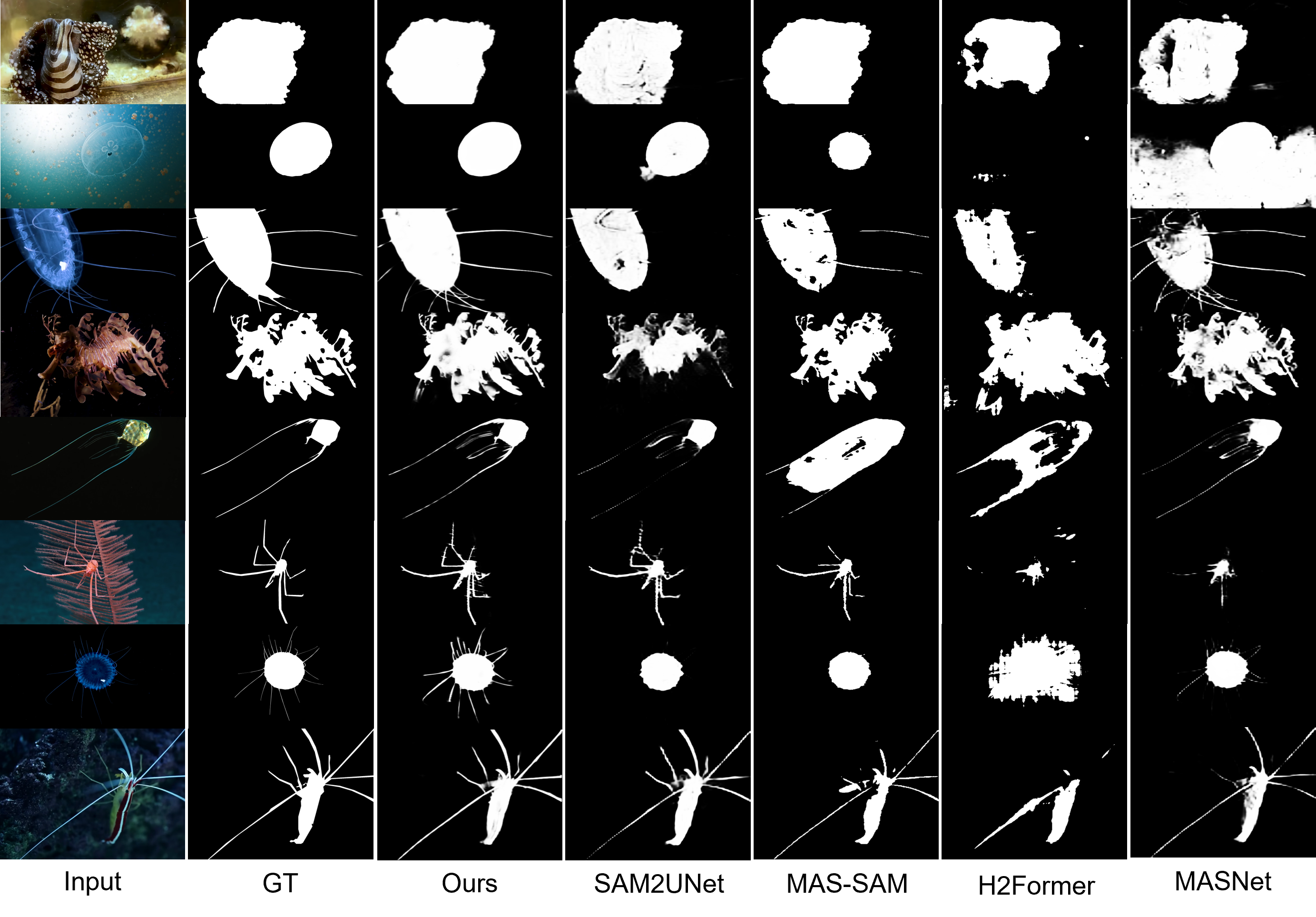}
	\caption{Qualitative comparison of segmentation results on challenging scenarios from the GBU-UCOD dataset. From left to right: Input Image, Ground Truth (GT), DeepTopo-Net (Ours), SAM2-Unet \cite{xiong2026sam2}, MAS-SAM \cite{wang2024massam}, H2Former \cite{he2023h2former}, and MASNet \cite{10113781}. Our method consistently preserves the topological connectivity of slender appendages (top rows) and accurately segments transparent bodies (bottom rows) where other SOTA methods fail or produce fragmented masks.}
	\label{fig:visual}
\end{figure*}

\subsection{Performance Analysis}
\label{subsec:performance}

\noindent {\bf Quantitative analysis.}
Table \ref{tab:sota} summarizes the quantitative results of our DeepTopo-Net against 10 state-of-the-art competitors on three challenging underwater benchmarks under five evaluation metrics. It can be seen clearly that general COD methods \cite{fan2021concealed} perform sub-optimally on deep-sea tasks due to the severe optical degradation. Furthermore, our method achieves the best results in most comparisons, especially on the newly proposed GBU-UCOD dataset. This dataset represents the most difficult vertical zonation scenarios, yet DeepTopo-Net is significantly ahead of the second-best SAM2-UNet that the $S_\alpha$ indicator is higher by nearly 1\% and $F_\beta^w$ by 0.7\%, reflects the strong robustness of our method in handling extreme deep-sea environments. On the remaining two datasets MAS3K and RMAS, our DeepTopo-Net has also achieved consistent leadership in terms of structural accuracy ($S_\alpha$) and boundary quality ($MAE$).

\noindent {\bf Qualitative analysis.}
Fig. \ref{fig:visual} presents a visual comparison of DeepTopo-Net with representative competitors across typical deep-sea scenarios. It can be seen that when the target possesses extremely slender appendages, other methods like MASNet, H2Former, and MAS-SAM often fail to preserve topological connectivity, resulting in severely fragmented or discontinuous masks. Even the powerful SAM2-UNet tends to over-smooth these fine structures or produce disconnected segments, but our method successfully ``stitches'' these thin limbs, providing topologically complete predictions. Furthermore, for transparent or low-contrast targets (e.g., jellyfish in Row 2 and Row 5), competitors frequently miss the subtle boundaries or make completely erroneous shape predictions. In contrast, DeepTopo-Net accurately delineates the full body shape and delicate tentacles, demonstrating superior robustness against deep-sea optical degradation.

\noindent {\bf Sensitivity to Loss Weight $\lambda$.} 
We further investigate the impact of the balancing hyperparameter $\lambda$ in Eq. (10), as shown in Table \ref{tab:lambda_ablation}. Comparing the results with different $\lambda$ values, it can be seen that the performance initially improves and then drops. Setting $\lambda=0$ (no auxiliary supervision) yields suboptimal results. Increasing $\lambda$ to 0.1 achieves the best trade-off, where $S_\alpha$ peaks at 0.918. However, when $\lambda$ becomes too large (e.g., 0.5), the performance degrades significantly. This suggests that while auxiliary supervision helps learn robust features, an excessive weight distracts the encoder from the primary segmentation objective.

\begin{table}[h]
\centering
\caption{Sensitivity analysis of the loss weight $\lambda$ on GBU-UCOD.}
\label{tab:lambda_ablation}
\begin{tabular}{l|cccc}
\hline
\textbf{Weight} $\lambda$ & $S_\alpha\uparrow$ & $F_\beta^w\uparrow$ & $mE_\phi\uparrow$ & MAE$\downarrow$ \\ \hline
0 (No Aux) & 0.908 & 0.845 & 0.938 & 0.030 \\
0.05       & 0.914 & 0.852 & 0.938 & 0.029 \\
\textbf{0.1 (Ours)} & \textbf{0.918} & \textbf{0.859} & \textbf{0.939} & \textbf{0.028} \\
0.2        & 0.912 & 0.850 & 0.936 & 0.030 \\
0.5        & 0.901 & 0.835 & 0.929 & 0.033 \\ \hline
\end{tabular}
\end{table}

\subsection{Ablation Study}
\label{subsec:ablation}

\noindent {\bf Effectiveness of WCAP.} Comparing the first and second rows in Table \ref{tab:ablation_all}, it can be seen that the addition of WCAP effectively enhances the network's performance. The improvement is most significant on GBU-UCOD, where mIoU increases by 2.3\%. This proves that WCAP effectively handles non-uniform optical degradation in deep-sea environments.

\noindent {\bf Effectiveness of ATRM.} Comparing the first and third rows in Table \ref{tab:ablation_all}, it can be seen that the addition of ATRM effectively improves structural metrics across all datasets. For instance, $F_\beta^w$ on RMAS increases significantly. This proves that ATRM effectively stitches fragmented limbs and maintains topological connectivity for marine organisms.

\noindent {\bf Synergistic Effectiveness.} Moreover, from the comparison between the intermediate rows and the last row in Table \ref{tab:ablation_all}, it can be seen that the combination of WCAP and ATRM achieves the best results. WCAP provides clearer features for ATRM to refine, leading to a peak mIoU of 0.829 on GBU-UCOD, which outperforms the baseline by 4.4\%.

\begin{table*}[t]
\centering
\caption{Ablation study of different components on MAS3K, RMAS, and GBU-UCOD Dataset. \textbf{B}: Baseline (MAE-based), \textbf{WCAP}: Water-Conditioned Adaptive Perceptor, \textbf{ATRM}: Abyssal-Topology Refinement Module. \textbf{mIoU} is included to evaluate region-level accuracy.}
\label{tab:ablation_all}
\resizebox{\textwidth}{!}{
\begin{tabular}{l|ccccc|ccccc|ccccc}
\hline
\multirow{2}{*}{\textbf{Configuration}} & \multicolumn{5}{c|}{\textbf{MAS3K}} & \multicolumn{5}{c|}{\textbf{RMAS}} & \multicolumn{5}{c}{\textbf{GBU-UCOD}} \\ \cline{2-16} 
 & mIoU$\uparrow$ & $S_\alpha\uparrow$ & $F_\beta^w\uparrow$ & $mE_\phi\uparrow$ & MAE$\downarrow$ & mIoU$\uparrow$ & $S_\alpha\uparrow$ & $F_\beta^w\uparrow$ & $mE_\phi\uparrow$ & MAE$\downarrow$ & mIoU$\uparrow$ & $S_\alpha\uparrow$ & $F_\beta^w\uparrow$ & $mE_\phi\uparrow$ & MAE$\downarrow$ \\ \hline
Baseline (B) & 0.770 & 0.865 & 0.812 & 0.915 & 0.032 & 0.708 & 0.832 & 0.775 & 0.908 & 0.035 & 0.785 & 0.870 & 0.810 & 0.932 & 0.036 \\
B + WCAP & 0.785 & 0.886 & 0.838 & 0.928 & 0.027 & 0.722 & 0.854 & 0.802 & 0.924 & 0.028 & 0.808 & 0.894 & 0.832 & 0.936 & 0.032 \\
B + ATRM & 0.792 & 0.892 & 0.849 & 0.934 & 0.024 & 0.728 & 0.861 & 0.811 & 0.929 & 0.026 & 0.815 & 0.901 & 0.840 & 0.937 & 0.031 \\
\textbf{Ours} & \textbf{0.804} & \textbf{0.905} & \textbf{0.864} & \textbf{0.941} & \textbf{0.021} & \textbf{0.742} & \textbf{0.875} & \textbf{0.826} & \textbf{0.939} & \textbf{0.021} & \textbf{0.829} & \textbf{0.918} & \textbf{0.859} & \textbf{0.939} & \textbf{0.028} \\ \hline
\end{tabular}
}
\end{table*}

\section{Conclusion}
\label{sec:conclusion}

In this paper, we propose DeepTopo-Net, a physically-informed framework that achieves SOTA performance in deep-sea object detection. Unlike general COD methods that rely on Euclidean space assumptions and struggle with non-uniform light propagation, our approach explicitly addresses optical degradation and topological fragmentation. By integrating the WCAP and ATRM modules, we redefine feature sampling within a Riemannian manifold to effectively stitch slender limbs that are typically invisible to conventional architectures. Extensive experiments on the GBU-UCOD dataset validate the adaptability of our method across various vertical zones. We plan to explore multi-modal fusion in future work to further enhance robustness in complex seabed environments.

\section*{Declaration of competing interest}
The authors declare that they have no known competing financial interests or personal relationships that could have appeared to influence the work reported in this paper.

\

\printcredits

\bibliographystyle{cas-model2-names}

\bibliography{cas-refs}



\end{document}